\documentclass[wcp]{jmlr}

\usepackage{longtable}

\usepackage{booktabs}

\usepackage{lineno}

\makeatletter
\let\Ginclude@graphics\@org@Ginclude@graphics 
\makeatother

\jmlrvolume{222}
\jmlryear{2023}
\jmlrworkshop{ACML 2023}

\title[C2LSE]{Active Level Set Estimation for Continuous Search Space with Theoretical Guarantee}

\author{\Name{Giang Ngo}* \Email{g.ngo@deakin.edu.au}\\
        \Name{Dang Nguyen} \Email{d.nguyen@deakin.edu.au}\\
        \Name{Dat Phan-Trong} \Email{trongp@deakin.edu.au}\\
        \Name{Sunil Gupta} \Email{sunil.gupta@deakin.edu.au}\\
        \addr Applied Artificial Intelligence Institute (A2I2), Geelong, Victoria, Australia\\
        \addr \textbf{*} Corresponding Author}
\editors{Berrin Yan{\i}ko\u{g}lu and Wray Buntine}

\begin{document}

\maketitle

\begin{abstract}
A common problem encountered in many real-world applications is level set estimation where the goal is to determine the region in the function domain where the function is above or below a given threshold. When the function is black-box and expensive to evaluate, the level sets need to be found in a minimum set of function evaluations. Existing methods often assume a discrete search space with a finite set of data points for function evaluations and estimating the level sets. When applied to a continuous search space, these methods often need to first discretize the space which leads to poor results while needing high computational time. While some methods cater for the continuous setting, they still lack a proper guarantee for theoretical convergence. To address this problem, we propose a novel algorithm that does not need any discretization and can directly work in continuous search spaces. Our method suggests points by constructing an acquisition function that is defined as a measure of confidence of the function being higher or lower than the given threshold. A theoretical analysis for the convergence of the algorithm to an accurate solution is provided. On multiple synthetic and real-world datasets, our algorithm successfully outperforms state-of-the-art methods.
\end{abstract}

\section{Introduction}
In many real-world problems, it is necessary to find a region in a search space \(S\) of a function \(f\) such that \(f\) is above/below a certain threshold of interest. Defined as the \textit{level set estimation} (LSE) problem (\cite{activelse}), it becomes challenging when the function is black-box and its evaluation is expensive. An example of such a problem is the design process of vehicle structure in which the goal is to find the designs that are safe, which is ensured by a Head Injury Criterion (HIC) being lower than a predefined threshold. LS-DYNA, a multi-physics simulator (\cite{hallquist1998livermore}), is run to find a set of designs that meet the above criteria before the actual vehicles are built and tested. Both running the simulator and actual vehicle tests are expensive, so it is important to sample-efficiently find a set of designs in advance such that they have a higher chance of satisfying the HIC threshold. Another example of the LSE problem is algorithmic assurance for machine learning (\cite{NEURIPS2018_cc709032}), where the goal is to find the operating scenarios (e.g. the brightness of input images for image classification or the camera angle for object detection) where a pre-trained machine learning algorithm will retain its performance above a user-specified accuracy level. Since testing the machine learning algorithm with a large set of scenarios is computationally expensive, it is important to accurately obtain the level sets with as few evaluations as possible.

\noindent\textbf{Related Works}

Previous works on sample-efficient LSE problems use an active learning scheme where the information about \(f\) is acquired by sequentially querying it at a set of points in the search space (\cite{straddle, activelse, truvar, rmile}). To accurately and quickly determine the level set, both exploring the search space and exploiting data points near the threshold boundary are required (\cite{rmile}). To achieve this goal, most algorithms propose optimizing an acquisition function that balances both factors to suggest the next query point. The LSE algorithm (\cite{activelse}), TruVar (\cite{truvar}), Straddle heuristic (\cite{straddle}), and RMILE (\cite{rmile}) are among the most well-known algorithms for the active LSE problem. Although the search space often involves multiple continuous variables, most of the methods estimate the level sets only for a finite discretization of the search space through an online classification scheme.

The LSE algorithm in \cite{activelse} utilized confidence intervals to perform online classification for a finite subset of data points. Its acquisition function considers the uncertainty at a point \(x\) about whether it is in the superlevel (i.e. function value being higher than the given threshold) or sublevel (i.e. function value being lower than the given threshold) sets. Search space warping was used in \cite{warping} to improve the performance of the LSE algorithm. Another online classification algorithm is TruVar (\cite{truvar}) which was proposed to unite both Bayesian optimization and the LSE problem under point-wise cost and heteroscedastic noise settings. Both the LSE algorithm and TruVar were related to the GP-UCB algorithm (\cite{srinivas}) which is a popular acquisition function for Bayesian optimization. While these level set methods are well analyzed and come with some form of convergence guarantees, \emph{they are only designed for finite search spaces and cannot handle continuous search spaces, which is often required in many practical applications}. When forced to work with continuous search spaces, these methods use a discretization, and the accuracy of these methods depends on the granularity of the discretization, and given a black-box function, it is difficult to set the correct granularity in advance. In addition, although existing methods may achieve accurate level set estimation by operating on a very densely discretized search space, such a discretization substantially increases the computational complexity as the acquisition function needs to be evaluated at all unclassified points on the grid.

The active LSE methods that are designed to directly work in continuous spaces are limited. The Straddle heuristic (\cite{straddle}), one of the earliest LSE works, employs Gaussian Process to construct an acquisition function on the continuous search space with a goal to perform global exploration and sampling close to the level threshold boundary. While this algorithm has been shown to work in practice, there is no theoretical understanding of its convergence or its sample efficiency. A recent algorithm (RMILE) utilizing expected improvement was proposed in \cite{rmile}. RMILE can be used for continuous spaces, however, its convergence behavior has only been analyzed for a finite grid. Further, due to the need for one step look-ahead, this algorithm is computationally very expensive. Recent extensions of RMILE to incorporate input uncertainty (\cite{9252941}), cost dependency (\cite{9272630}), and uncontrollable environmental variables (\cite{inatsu2021active}) suffer from the same drawback. Another work that was proposed for level set estimation in high dimensional spaces uses a Bayesian Neural Network (\cite{HaHuong2021HDLS}), which is also capable of operating on continuous search spaces. However, this method does not come with any analysis of its convergence or sample efficiency. \cite{HaHuong2021HDLS} also acknowledged a worse performance on low-dimensional search spaces compared to GP-based approaches.

Thus, designing an active LSE method that can work in continuous spaces and has a theoretical guarantee of its convergence and sample efficiency is an \emph{open problem}. To address this problem, we propose \textit{Confidence-based Continuous Level Set Estimation} (C2LSE), a novel algorithm that directly performs level set estimation on a continuous search space by improving the confidence of classifying points into the superlevel/sublevel sets while also balancing the exploration-exploitation trade-off. At each iteration, our algorithm queries a point where the difference of probabilities of the function being higher and lower than the threshold level is minimum. This is done through a simple yet effective design of our acquisition function. Our algorithm does not rely on the density of any finite discretization, which allows for better classification performance compared to those that operate on such discretized spaces. We also theoretically analyze our algorithm and provide proof of its convergence and its sample efficiency for the active LSE problem. We perform experiments on multiple level set tasks to demonstrate the competitive performance of our method compared to the state-of-the-art baselines.

In summary, the contributions of our work are as follows:
\begin{itemize}
    \item We propose C2LSE, a novel algorithm for actively estimating the level set of a function in continuous search spaces.
    \item We theoretically analyze our algorithm and provide a proof of its convergence and its sample efficiency.
    \item We apply C2LSE on many LSE tasks covering both synthetic and real-world applications, and show superior performance over the current state-of-the-art methods.
\end{itemize}

\section{Background}
When the number of available data is limited with respect to the whole search space \(S\), Gaussian Process (GP) is often used to model the unknown function \(f\) due to the fact that such a nonparametric probabilistic model can give both prediction and uncertainty about \(f\) at any point in \(S\).

\noindent\textbf{Gaussian Process}
A GP prior assumed over \(f\) is defined by its mean function \(\mu(\textbf{x})\) and its kernel \(k(\textbf{x},\textbf{x}^\prime)\). Given \(t\) noisy observations \(\textbf{y}_t=[y_1,...,y_t]^T\) of \(f\) at \(\{\textbf{x}_i\}_{i=1}^t\) where \(y_i=f(\textbf{x}_i)+\eta_i\) and \(\eta_i\sim \mathcal{N}(0,\sigma^2)\) for \(i=1,...,t\), the posterior distribution of \(f\) is also a GP whose mean and variance can be computed as follows:
\begin{align}
    \mu_t(\textbf{x})&=\textbf{k}_t(\textbf{x})^T(\textbf{K}_t+\sigma^2I)^{-1}\textbf{y}_t\nonumber\\
    \sigma_t^2(\textbf{x})&=k(\textbf{x},\textbf{x})-\textbf{k}_t(\textbf{x})^T(K_t+\sigma^2I)^{-1}\textbf{k}_t(\textbf{x})\nonumber
\end{align}
where \(\textbf{k}_t(\textbf{x})=[k(\textbf{x}_i,\textbf{x})]_{i=1}^t\), \(\textbf{K}_t=[k(\textbf{x}_i,\textbf{x}_j)]_{i,j=1}^t\). At each iteration, the posterior mean \(\mu_t(\textbf{x})\) gives an estimate of \(f(\textbf{x})\) while the posterior variance \(\sigma_t^2(\textbf{x})\) indicates the uncertainty about the estimate.
\section{Proposed Method}
\subsection{Problem Statement}
We first state the problem of level set estimation in an active learning setting as follows:

\noindent\textbf{Active Level Set Estimation Problem}: Given a black-box function \(f: S\rightarrow R\) with \(S\subset R^d\), level set estimation problem aims to classify any point \(\textbf{x}\in S\) into two sets: a superlevel set \(H=\{\textbf{x}\in S|f(\textbf{x})> h\}\) and a sublevel set \(L=\{\textbf{x}\in S|f(\textbf{x}) < h\}\), where \(h \in R\) is a given threshold level.

A common way to solve the LSE problem is to query \(f\) multiple times to obtain the function values and then build a model (e.g. a GP) based on the queried points to approximate \(f\). In an active learning setting, the aim is to query \(f\) as few times as possible.

Assuming a GP prior over the given black-box function \(f\), our algorithm is motivated by the idea of sampling where the GP is the least confident in classifying \(\textbf{x}\). We provide below a definition of level set classification confidence using GP.

\begin{definition}
    \label{def:1}
    Suppose that \(f\) is approximated by a posterior \textup{GP} where the posterior mean and variance of \textup{GP} at \(\mathbf{x}\in S\) is given by \(\mu_{\textup{GP}}(\mathbf{x})\) and \(\sigma_{\textup{GP}}^2(\mathbf{x})\) respectively. The level set classification confidence of a point \(\mathbf{x}\in S\) is \(C_{\textup{GP}}(\mathbf{x})=|P(f(\mathbf{x})>h)-P(f(\mathbf{x})<h)|\).
\end{definition}

With \(\Phi\) being the standard normal CDF, we know that:
\begin{align}
    C_{\textup{GP}}(\textbf{x})&=2\Phi(\frac{|\mu_{\textup{GP}}(\textbf{x})-h|}{\sigma_{\textup{GP}}(\textbf{x})})-1\nonumber\\
    &=2\Phi(\frac{1}{Z_{\textup{GP}}(\textbf{x})})-1 \text{ with } Z_{\textup{GP}}(\textbf{x})=\frac{\sigma_{\textup{GP}}(\textbf{x})}{|\mu_{\textup{GP}}(\textbf{x})-h|}\nonumber
\end{align}

The point at which \(C_{\textup{GP}}(\textbf{x})\) is the smallest is where the GP is the least confident for level set estimation. Since \(\Phi\) is monotonically increasing, it is also the point that maximizes \(Z_{\textup{GP}}(\textbf{x})\). In practice, \(Z_{\textup{GP}}(\textbf{x})\) cannot be used directly as an acquisition function because the algorithm may get ``stuck" to points where the function value is extremely close to \(h\) (i.e. where \(C_{\textup{GP}}(\textbf{x})\approx0\)) even though the GP uncertainty \(\sigma_{\textup{GP}}(\textbf{x})\) is already small at those points. This leads to poor exploration behavior for the algorithm. To avoid this problem, we use a small hyperparameter \(\epsilon>0\) in our acquisition function which allows the algorithm to continue its explorations. Our acquisition function at iteration \(t\) is as follows:
\begin{align}\label{acq}
    a_t(\textbf{x})=\frac{\sigma_{t-1}(\textbf{x})}{\text{max}(\epsilon, |\mu_{t-1}(\textbf{x})-h|)}
\end{align}
with \(\epsilon\) being a small positive hyperparameter and \(\mu_{t-1}(\textbf{x})\) and \(\sigma_{t-1}^2(\textbf{x})\) being the posterior mean and variance respectively given by the GP that models \(f\) after \(t-1\) iterations. With this acquisition function, the next query point is then \(\textbf{x}_t=\text{argmax}_{\textbf{x}\in S}(a_t(\textbf{x}))\). We account for \(\epsilon\) in our theoretical analysis and show how this parameter influences the convergence of our algorithm. 

We refer to our method as Confidence-based Continuous Level Set Estimation (C2LSE), summarized in Algorithm \ref{alg:one}.

\begin{algorithm2e}
  \caption{C2LSE for Active Level Set Estimation}\label{alg:one}
  \KwIn{GP prior (\(\mu_0, k, \sigma_0\)), threshold \(h\), budget \(T\), exploration parameter \(\epsilon\), parameter \(\beta\).}
  \(D_0\leftarrow\emptyset; t\leftarrow1\)\;
  \While{\(t<T\)}{
    \(\textbf{x}_t=\text{argmax}_{\textbf{x}\in S}(a_t(\textbf{x}))\)\;
    Observe \(y_{t}=f(\textbf{x}_t)+\eta_i\)\;
    \(D_{t}=D_{t-1}\cup(\textbf{x}_t, y_t)\)\;
    Update GP with \(D_{t}\)\;
    \(t=t+1\);
  }
  \KwOut{
    \\
    \(\hat{H}=\{\textbf{x}\in S:\mu_T(\textbf{x})-\beta\sigma_T(\textbf{x})>h\}\)\;
    \(\hat{L}=\{\textbf{x}\in S:\mu_T(\textbf{x})+\beta\sigma_T(\textbf{x})< h\}\)\;
    \(\hat{M}=\{\textbf{x}\in S:\mu_T(\textbf{x})-\beta\sigma_T(\textbf{x})\leq h\leq\mu_T(\textbf{x})+\beta\sigma_T(\textbf{x})\}\)\;
  }
\end{algorithm2e}

The main goal of the LSE problem is being able to classify \(x\in S\) into the correct level set. For this purpose, the classification rule is applied after running the algorithm for the given budget of \(T\) iterations. Any point \(x\in S\) can be classified into the predicted superlevel set \(\hat{H}\), sublevel set \(\hat{L}\), and unclassified set \(\hat{M}\). However, online classification with the same rule can also be employed after each iteration although a certain number of iterations are required for confident and accurate classification as analyzed in the following section. We will also show in the next section that the unclassified points in \(\hat{M}\) will be limited to points close to the threshold boundary given a sufficient budget.

\section{Theoretical Analysis}
In this section, we present a theoretical analysis of our algorithm C2LSE and provide a lower bound on the number of iterations needed for C2LSE to be able to classify a data point into the superlevel/sublevel set with a given level of confidence.

\subsection{Preliminaries}
The information gain about \(f\) after observing \(\textbf{y}_t\) (\cite{10.5555/1146355}) plays a crucial role in analyzing the convergence of C2LSE.

\begin{definition}
    \label{def:2}
    After \(t\) noisy observations \(\mathbf{y}_t\), the information gain about \(f\) is:
    \begin{center}
        \(I(\mathbf{y}_t,f)=H(\mathbf{y}_t)-H(\mathbf{y}_t|f)\)
    \end{center}
\end{definition}

\begin{lemma}
    \label{lem:2}
    For a \textup{GP}, the information gain \(I(y_{1:t},f)\) has the following lower bound:
    \begin{align}
        I(\mathbf{y}_t,f)\geq \frac{\textup{log}(1+\sigma^{-2})}{2}\sum_{t=1}^{T}\sigma^2_{t-1}(\mathbf{x}_t)\label{inq:infg}
    \end{align}
    \begin{proof}
    The proof is similar to Lemma 5.4 in \cite{srinivas}.
    \end{proof}
\end{lemma}

\begin{definition}
    \label{def:3}
    The maximum information gain \(\gamma_t\) over all possible sets of \(t\) noisy observations is:
    \begin{align}
        \gamma_t=\underset{\mathbf{y}_t}{\textup{max}}I(\mathbf{y}_t,f)\label{eq:maxinfg}
    \end{align}
\end{definition}

\subsection{Theoretical results}
As mentioned in section 3, C2LSE's acquisition function \(a_t(\textbf{x})\) is indicative of the GP's confidence in classifying the point \(\textbf{x}\) into the superlevel set or the sublevel set. Thus, the analysis will focus on showing the number of iterations needed for the acquisition function to reach a threshold that is connected to the confident classification of any \(\textbf{x}\). We will first show an upper bound for the value of \(a_t(\textbf{x}_t)\) using information-theoretic terms,  and devise the number of iterations needed to reach a certain level of classification confidence. In addition, we show that C2LSE also achieves highly accurate classification by connecting the lower bound of the classification confidence with the definition of accuracy in an LSE problem.

First, an upper bound for the average acquisition function's value is established as follows:

\begin{lemma}
    \label{lem:3}
    Let \(\mathbf{x}_t\) (where \(1\leq t\leq T\)) be the data point selected by C2LSE at iteration \(t\). With \(C_1=\frac{2}{\textup{log}(1+\sigma^{-2})}\), it holds that:
    \begin{equation}
        \frac{1}{T}\sum_{t=1}^{T}a_t(\mathbf{x}_t)\leq \sqrt{\frac{C_1\gamma_T}{T\epsilon^2}},
    \end{equation}
    \begin{proof}
        For any \(T>0\), we know that:
        \begin{center}
            \begin{align*}
                TC_1\gamma_T & \geq TC_1I(\mathbf{y}_t,f)    &\text{(using Eq. \ref{eq:maxinfg})}                                        \\
                             & \geq T\sum_{t=1}^{T}\sigma^2_{t-1}(\mathbf{x}_t)     &\text{(using Ineq. \ref{inq:infg})}                         \\
                             & \geq (\sum_{t=1}^{T}\sigma_{t-1}(\mathbf{x}_t))^2    &\text{(by Cauchy-Schwarz)}\\
                             & = (\sum_{t=1}^{T}a_t(\mathbf{x}_t)\textup{max}(\epsilon,|\mu_{t-1}(\mathbf{x})-h|))^2     &\text{(by Eq. \ref{acq})}\\
                             & \geq \epsilon^2(\sum_{t=1}^{T}a_t(\mathbf{x}_t))^2&
            \end{align*}
        \end{center}
        This lemma holds as a result.
    \end{proof}
\end{lemma}

As the algorithm becomes more confident in classifying the level sets with more iterations, its convergence rate is given below:

\begin{theorem}
    \label{thm:1}
     Using a Gaussian Process with a smooth kernel, C2LSE can classify any point \(\mathbf{x}\) in the search space with a confidence of at least \(2\Phi(\beta)-1\) if it is run for at least T iterations such that:
    \begin{align}
        \frac{T}{\gamma_T}\geq \frac{\beta^2C_1}{\epsilon^2} \label{inq:rate}
    \end{align}
    and that 
    \begin{align}
        a_T(\mathbf{x}_T)\leq a_t(\mathbf{x}_t), \forall t\leq T \label{inq:rate2}
    \end{align}
    \begin{proof}
        By definition:
        \begin{center}
            \begin{align*}
                a_t(\textbf{x})=\frac{\sigma_{t-1}(\textbf{x})}{\textup{max}(\epsilon,|\mu_{t-1}(\textbf{x})-h|)}\leq\frac{\sigma_{t-1}(\textbf{x})}{\epsilon}\\
            \end{align*}
        \end{center}
        Given a smooth kernel, $\sigma^2_{t-1}(\mathbf{x})$ will be a Lipschitz continuous function, so $|\sigma_{t-1}(\mathbf{x})-\sigma_{t-1}(\mathbf{x}')|\leq L||\mathbf{x}-\mathbf{x}'||_1$ $\forall\mathbf{x},\mathbf{x}'\in S$. Using the idea of covering numbers (e.g. Lemma 5.7 in Srinivas et al. (2010)), we can discretize the continuous search space $S$ as $S_t$ at iteration $t$ such that $||\mathbf{x}-[\mathbf{x}]_t||_1\leq\frac{c}{t^2}$ for a constant $c$ with $[\mathbf{x}]_t$ being the closest point to $\mathbf{x}$ in $S_t$. This means $|\sigma_{t-1}(\mathbf{x})-\sigma([\mathbf{x}]_t)|\leq L||\mathbf{x}-[\mathbf{x}]_t||_1=O(\frac{1}{t^2})$. Our algorithm will visit all points in any arbitrarily dense finite grid as visiting only a subset of the grid will eventually lead to zero acquisition value (due to zero variance) for points within the subset and points outside the subset will be preferred. This means $\sigma_{t-1}([\mathbf{x}])_t$ tends to zero as $t\to\infty$, so we can say $\lim_{t\to\infty}\frac{\sigma_{t-1}(\textbf{x}_t)}{\epsilon}=0$. Since \(a_t(\textbf{x}_t)\geq0\) \(\forall t\), it follows that \(\lim_{t\to\infty}a_t(\textbf{x}_t)=0\) by Squeeze Theorem.
        
        For any smooth kernel employed by the GP prior of C2LSE, without loss of generality we have \(\sigma_0(\textbf{x}_1)=1\), which means \(a_1(\textbf{x}_1)>0\) and consequently \(a_1(\textbf{x}_1)>\lim_{t\to\infty}a_t(\textbf{x}_t)=0\). As the sequence of \(\{a_t(\textbf{x}_t)\}_{t\geq1}\) converges to 0 from a positive number with increasing \(t\), there exists an iteration $T$ such that \(a_T(\textbf{x}_T)\leq a_t(\textbf{x}_t)\) \(\forall t\leq T\), which results in:
        \begin{align}
            a_T(\textbf{x}_T) &\leq \frac{1}{T}\sum_{t=1}^{T}a_t(\textbf{x}_t) \label{inq:avg}\\
                              &\leq \sqrt{\frac{C_1\gamma_T}{T\epsilon^2}}\nonumber
        \end{align}
        If \(T\) also satisfies Ineq. \ref{inq:rate}, we have \(a_T(\textbf{x}_T)\leq \frac{1}{\beta}\) and also \(a_T(\textbf{x})\leq \frac{1}{\beta}\) \(\forall \textbf{x}\in S\). Since \(C_T(\textbf{x})\approx 2\Phi(\frac{1}{a_T(\textbf{x})})-1\) for small \(\epsilon\), this leads to \(C_T(\textbf{x})\geq 2\Phi(\beta)-1\). The theorem holds as a result.
    \end{proof}
\end{theorem}

\begin{remark}
Given that \(\{a_t(\mathbf{x}_t)\}_{t\geq1}\) is a positive sequence converging to 0, Ineq. \ref{inq:rate2} is always satisfied for increasing \(T\), and since Ineq. \ref{inq:rate} will eventually be met, C2LSE is guaranteed to converge while the convergence rate is determined by the two conditions in Ineq. \ref{inq:rate} and Ineq. \ref{inq:rate2}. In fact, Ineq. \ref{inq:rate2} is an overly strict condition because most positive sequences converging to 0 like \(\{a_t(\mathbf{x}_t)\}_{t\geq1}\) will satisfy it with increasing \(T\). Moreover, Ineq. \ref{inq:rate2} is the worst-case condition with the goal to satisfy Ineq. \ref{inq:avg}, so there exists \(T\) such that Ineq. \ref{inq:avg} is satisfied but Ineq. \ref{inq:rate2} is not (i.e. it is not necessary for the last item of a sequence to be the smallest in order for the last item to be smaller than the sequence average). Therefore, the convergence rate of C2LSE is usually quantified by Ineq. \ref{inq:rate}.
\end{remark}

Theorem \ref{thm:1} shows that C2LSE can confidently classify any point with a high probability in a continuous search space given sufficient iterations. Next, we will show that these highly confident classifications are also correct in a similar manner compared to the definition of \(\epsilon\)-accuracy introduced in \cite{truvar}.

\begin{definition}
    \label{def:4}
     A solution of a level set estimation problem with \(f, S, h\) is \(\epsilon\)-accurate if it can classify any point \(\mathbf{x}\) into \(\hat{H}, \hat{L}\) and \(\hat{M}\) such that:
    \begin{itemize}
        \item \(f(\mathbf{x})>h\) if \(\mathbf{x}\in \hat{H}\)
        \item \(f(\mathbf{x})<h\) if \(\mathbf{x}\in \hat{L}\)
        \item \(|f(\mathbf{x})-h|\leq\epsilon\) if \(\mathbf{x}\in \hat{M}\)
    \end{itemize}
\end{definition}

Note that the above is defined on the whole continuous search space while \(\epsilon\)-accuracy in \cite{truvar} is defined only for a finite set of data points. We will show that C2LSE guarantees a \(2\epsilon\)-accurate solution as follows:

\begin{theorem}
    \label{thm:2}
     If C2LSE is run for at least \(T\) iterations as per Theorem \ref{thm:1}, it will reach a \(2\epsilon\)-accurate solution with probability at least \(2\Phi(\beta)-1\) for any point \(\mathbf{x}\).

    \begin{proof}
        Conditioned on \(f(\textbf{x}) \sim \mathcal{N}(\mu_{t-1}(\textbf{x}),\sigma_{t-1}(\textbf{x}))\), we know that 
        \begin{equation}
            f(\textbf{x})\in[\mu_{t-1}(\textbf{x})-\beta\sigma_{t-1}(\textbf{x}),\mu_{t-1}(\textbf{x})+\beta\sigma_{t-1}(\textbf{x})],\nonumber
        \end{equation}
        with probability \(2\Phi(\beta)-1\). Given our classification rule in Algorithm \ref{alg:one}, any point in either \(\hat{H}\) and \(\hat{L}\) satisfies the first and second requirements of Definition \ref{def:4} with probability at least \(2\Phi(\beta)-1\).
        
        When \(\frac{T}{\gamma_T}\geq \frac{\beta^2C_1}{\epsilon^2}\), we know that:
        \begin{align} \label{inq:beta}
            \frac{1}{\beta}     \geq a_T(\textbf{x}_T) &\geq a_T(\textbf{x}) = \frac{\sigma_{T-1}(\textbf{x})}{\textup{max}(\epsilon, |\mu_{T-1}(\textbf{x})-h|)}
        \end{align}
        
        The classification rule for \(\hat{M}\) is equivalent to having \(|\mu_{T-1}(\textbf{x})-h|\leq \beta\sigma_{T-1}(\textbf{x})\) for \(\textbf{x}\). From Ineq. \ref{inq:beta}, if \(|\mu_{T-1}(\textbf{x})-h|>\epsilon\), we then have \(|\mu_{T-1}(\textbf{x})-h|\geq \beta\sigma_{T-1}(\textbf{x})\) which contradicts the classification rule of \(\textbf{x}\) in \(\hat{M}\). Therefore, any point \(\textbf{x}\in\hat{M}\) satisfies \(|\mu_{T-1}(\textbf{x})-h|\leq\beta\sigma_{T-1}(\textbf{x})\leq\epsilon\). The result is that:
        \begin{align}
            h-\epsilon\leq \mu_{T-1}(\textbf{x})&\leq h+\epsilon\nonumber\\
            \Longrightarrow h-2\epsilon\leq\mu_{T-1}(\textbf{x})-\beta\sigma_{T-1}(\textbf{x})&<\mu_{T-1}(\textbf{x})+\beta\sigma_{T-1}(\textbf{x})\leq h+2\epsilon\nonumber\\
            \Longrightarrow |f(\textbf{x})-h|&\leq 2\epsilon\nonumber
        \end{align}
        with probability at least \(2\Phi(\beta)-1\).
        With all three conditions of a \(2\epsilon\)-accurate solution satisfied, the theorem holds.
    \end{proof}
\end{theorem}

Theorem \ref{thm:2} states that the points in \(\hat{H}\) and \(\hat{L}\) are classified correctly with high probability while those in \(\hat{M}\) are within \(2\epsilon\) from the threshold boundary with also a high probability. This analysis demonstrates the sample-efficiency of our method in acquiring accurate estimation of the level sets.

We note that our theoretical analysis applies to \emph{any kernel} in the Gaussian Process. The key term that depends on the kernel is $\gamma_T$ which has been related to $T$ in \cite{srinivas}. For popular kernels such as square-exponential and Matern, $\gamma_T$ is sublinear in $T$.
\section{Experimental Results}
This section studies the empirical performance of C2LSE on both synthetic functions and real-world applications compared to state-of-the-art algorithms.
\subsection{Experiment Setup}
\textbf{Baselines} C2LSE is compared against the state-of-the-art baselines: 
\begin{itemize}
    \item STR: the Straddle heuristic (\cite{straddle})
    \item LSE: the LSE algorithm (\cite{activelse})
    \item TRV: the TruVar algorithm (\cite{truvar})
    \item RMILE: the RMILE algorithm (\cite{rmile})
\end{itemize}
The classification accuracy of each algorithm is reported in terms of average F1-score with respect to the true superlevel/sublevel set. The experiments were randomly initialized and averaged over 10 runs. We note that full-grid ground truth data are for evaluation purposes only. Our algorithm can find the level sets in fewer number of function queries and directly works in continuous search spaces without requiring any grid.

\noindent\textbf{Synthetic Functions}

We use the following synthetic functions for the evaluation of C2LSE:

\begin{figure*}[t]
    \floatconts
    {fig:data}
    {\caption{The superlevel set (white) and sublevel set (gray) of two-dimensional data for evaluation}}
    {%
        \subfigure[MC2D]{%
            \includegraphics[width=4.4cm]{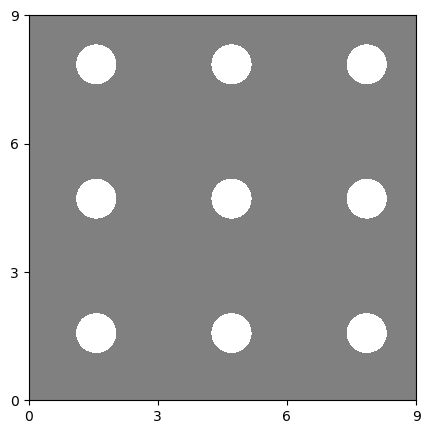}
            \label{fig:data:MC2D_data}
        }
        \subfigure[SIN2D]{%
            \label{fig:data:SIN2D_data}
            \includegraphics[width=4.4cm]{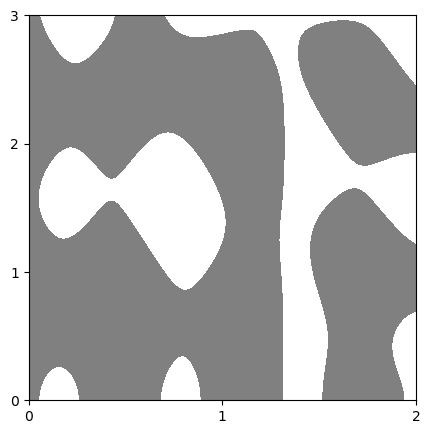}
        }
        \subfigure[CC2D]{%
            \label{fig:data:CC2D_data}
            \includegraphics[width=4.4cm]{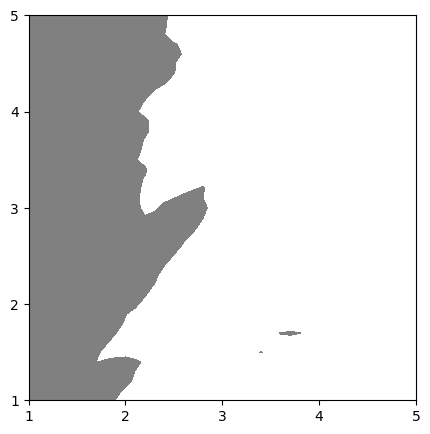}
        }
    }
\end{figure*}

\begin{itemize}
    \item MC2D (multi-circle 2D): \(f(\textbf{x})=e^{(\sin{x_1})^2\times(\sin{x_2})^2}\)
    
    The goal is to find the superlevel set at which \(f(\textbf{x})>2.2\) for \(x_1, x_2\in[0, 9]\). The ground-truth superlevel and sublevel sets are constructed by uniformly sampling a grid of \(100\times100\) data points.
    \item MC3D (multi-circle 3D): \(f(\textbf{x})=e^{(\sin{x_1})^2\times(\sin{x_2})^2\times(\sin{x_3})^2}\)
    
    The goal is to find the superlevel set at which \(f(\textbf{x})>1.6\) for \(x_1, x_2, x_3\in[0, 6]\). The ground-truth superlevel and sublevel sets are constructed by uniformly sampling a grid of \(30\times30\times30\) data points.
    \item SIN2D (Sinusoidal 2D): \(f(\textbf{x})=\sin(10x_1) + \cos(4x_2) - \cos(3x_1x_2)\)

    The goal is to find the superlevel set at which \(f(\textbf{x})>0.5\) for \(x_1\in[0, 2]\) and \(x_2\in[0, 3]\). The ground-truth superlevel and sublevel sets are constructed by uniformly sampling a grid of \(100\times100\) data points.
\end{itemize}

Besides SIN2D which is popularly used for evaluation in previous works (\cite{straddle, rmile}), MC2D and MC3D are introduced to challenge C2LSE and the baselines. Given appropriate thresholds, both contain multiple small and separated subregions in the superlevel set. This is considered harder to estimate the level sets than a search space where either the superlevel set or the sublevel set is dominant and spans over a largely connected subregion. Figure \ref{fig:data}a and Figure \ref{fig:data}b show the true level sets of MC2D and SIN2D respectively.

\noindent\textbf{Real-world Data}

Besides synthetic functions, C2LSE is also evaluated on three real-world datasets introduced below:
\begin{itemize}
    \item AA (Algorithmic Assurance): The algorithmic assurance problem, as introduced in \cite{NEURIPS2018_cc709032}, tries to efficiently discover scenarios where a machine learning algorithm deviates the most from its gold standard. Instead of finding the maximum deviation, we will find the scenarios where the performance of the machine learning algorithm stays above a given threshold.
    
    In particular, the goal of this experiment is to find the level of image distortions for which a LENET-5 (\cite{lenet}) trained on raw MNIST training digits has classification error less than 4\%. Number 9 is removed from the training and test data to avoid confusion with number 6 (due to rotation distortion involved). LENET-5 was trained for 20 epochs with a learning rate of \(10^{-3}\). Three image distortions are employed, namely scaling \(Sc\in[0.5, 1.5]\), rotation \(R\in[0, 340]\), and shear \(Sh_x, Sh_y\in[0, 20]\), in this exact order. A set of distortion \(\textbf{x}=(Sc, R, Sh_x, Sh_y)\) will be applied to all raw MNIST test digits where the trained LENET-5 will be evaluated on, which will result in the corresponding classification error. The step sizes for \((Sc, R, Sh_x, Sh_y)\) are 0.1, 20, 2, and 2 respectively, and the ground truth superlevel and sublevel sets contain 23,958 data points.
    \item CC2D (Car Crashworthiness 2D): This application of level set estimation for car design focuses on finding the set of design parameters of a car such that its Head injury criterion (HIC) is less than 250. There are two design parameters: \textit{tbumper} (mass of the front bumper) and \textit{thood} (mass of the front, hood, and underside of the bonnet). Both parameters range from 1 to 5. The ground truth superlevel and sublevel sets contain 1,681 data points as a 41\(\times\)41 grid of the two input parameters.
    \item CC6D (Car Crashworthiness 6D): This is another car design application with 6 design parameters: \textit{thood} (thickness of the hood), \textit{tgrill} (thickness of the grill), \textit{troof} (thickness of the roof), \textit{tbumper} (thickness of the bumper), \textit{trailf} (thickness of the front of rails) and \textit{trailb} (thickness of the back of rails). All input parameters range from 1 to 5, and form the ground truth superlevel and sublevel sets of 15,625 data points with a step size of 1 for each parameter. The goal is to find the set of input parameters for which the torsional mode frequency is less than 1.9Hz.
\end{itemize}
Figure \ref{fig:data}c shows the true level sets of CC2D.

\noindent\textbf{Parameter choice and kernel's hyperparameter tuning}
The parameter \(\epsilon\) is selected from \{0.01, 0.02, 0.05, 0.1, 0.2\}. All parameters/hyperparameters of the baselines are set according to the recommended values in the original works. Hyperparameters of a Matern-5 kernel (\cite{10.7551/mitpress/3206.001.0001}) are fitted via maximum likelihood estimation.

\subsection{Results and Discussion}

\begin{figure*}[t]
    \floatconts
    {fig:synthetic}
    {\caption{The level set classification performances of C2LSE and the baselines on synthetic functions.}}
    {%
        \subfigure[MC2D]{%
            \label{fig:MC2D}
            \includegraphics[width=4.85cm]{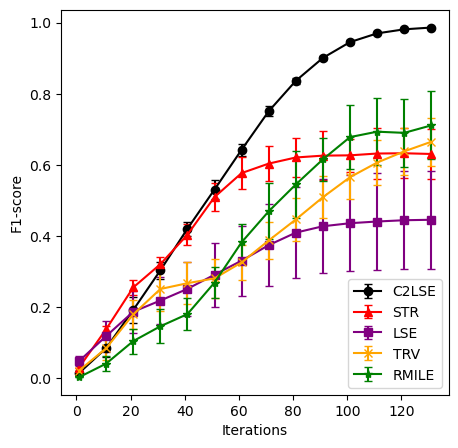}
        }
        \subfigure[MC3D]{%
            \label{fig:MC3D}
            \includegraphics[width=4.85cm]{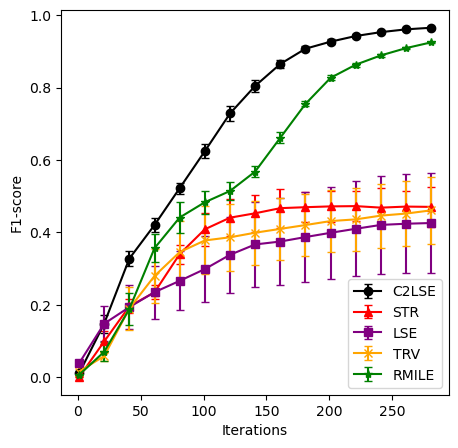}
        }
        \subfigure[SIN2D]{%
            \label{fig:SIN2D}
            \includegraphics[width=4.85cm]{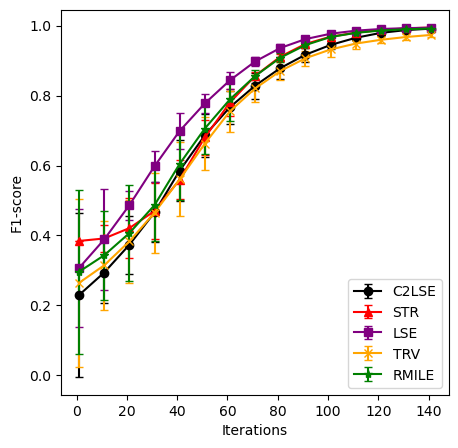}
        }
    }
\end{figure*}
\begin{figure*}[t]
    \floatconts
    {fig:realworld}
    {\caption{The level set classification performances of C2LSE and the baselines on real-world applications.}}
    {%
        \subfigure[AA]{%
            \label{fig:realworld:AA}
            \includegraphics[width=4.85cm]{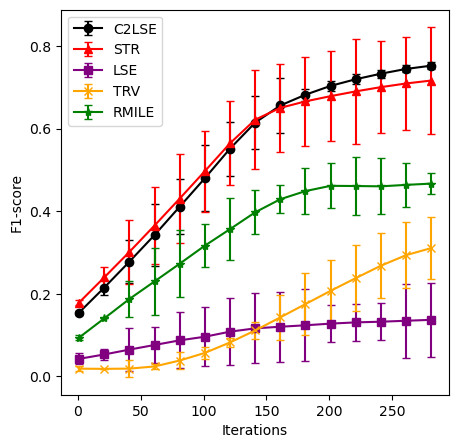}
        }
        \subfigure[CC2D]{%
            \label{fig:realworld:CC2D}
            \includegraphics[width=4.85cm]{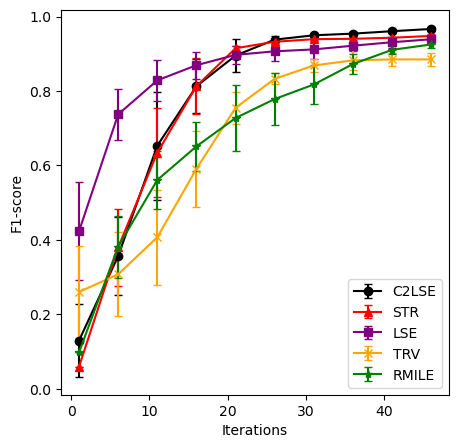}
        }
        \subfigure[CC6D]{%
            \label{fig:realworld:CC6D}
            \includegraphics[width=4.85cm]{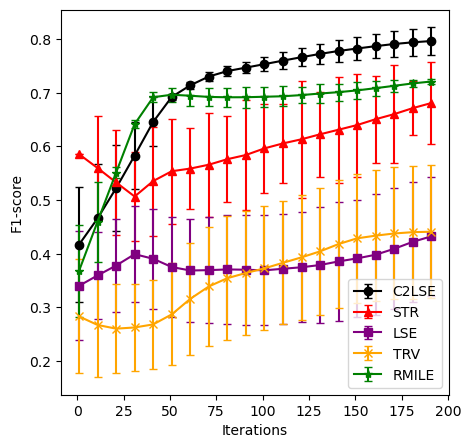}
        }
    }
\end{figure*}

\subsubsection{Main result on synthetic functions and real-world datasets}

Figure \ref{fig:synthetic} and Figure \ref{fig:realworld} show the classification performance of C2LSE and the baseline algorithms on synthetic and real-world datasets respectively. As seen, C2LSE consistently outperforms other algorithms across all six datasets with continuous search spaces including MC3D, AA, and CC6D which are higher dimensional. Other algorithms also have more varied classification performances compared to C2LSE due to their inconsistent global exploration among different runs as a result of those algorithms getting ``stuck'' at a discovered boundary region and failing to further discover other unknown regions.

In the MC2D experiment and somewhat in the MC3D experiment, C2LSE strictly dominates other algorithms by a significant margin. These are both difficult scenarios where the superlevel set is not one single large region (as is the case for CC2D) but scatters all over the search space. An additional challenge is \textit{data imbalance} when only a small fraction of the search space belongs to the superlevel set. In MC2D, the superlevel set accounts for approximately 7.8\% of the search space, and in MC3D, the ratio is about 7.5\%. On the other hand, C2LSE shows only a slight improvement over the baselines in CC2D and SIN2D. CC2D and SIN2D are less scattered than MC2D and MC3D and are also less imbalanced with 32.8\% and 31.52\% of the data in the targeted level sets respectively. The targeted level sets of both AA and CC6D are a contiguous range in the search space which also explains the smaller improvement. Thus, C2LSE holds the advantage in the challenging scenario of an LSE problem when multiple small and disconnected superlevel/sublevel sets exist in the search space which is harder to fully explore.

\subsubsection{Comparison with the LSE algorithm at different discretization levels}
Designed to work on finite search spaces, the classification performance of the LSE algorithm (\cite{activelse}) depends on the sparsity of the discretized search space. Figure \ref{fig:gridsize} shows the classification performance of C2LSE in comparison with the LSE algorithm on different sizes for MC2D, MC3D, and SIN2D. We note that for both C2LSE and the LSE algorithm, the evaluation is performed using the same level set ground truth as was used in Figure \ref{fig:synthetic}.

\begin{figure*}[t]
    \floatconts
    {fig:gridsize}
    {\caption{Comparing the classification performances of C2LSE with the LSE algorithm on \textit{different levels of discretization}.}}
    {%
        \subfigure[MC2D]{%
            \label{fig:gridsize:mc2d}
            \includegraphics[width=4.85cm]{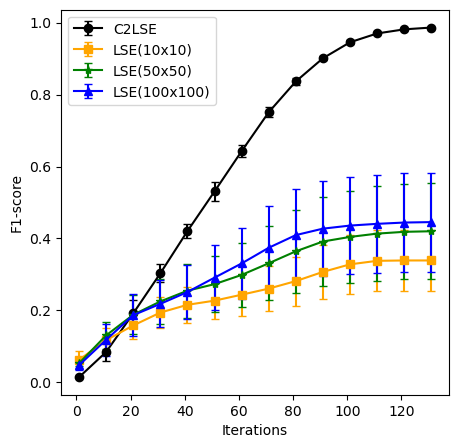}
        }\
        \subfigure[MC3D]{%
            \label{fig:gridsize:mc3d}
            \includegraphics[width=4.85cm]{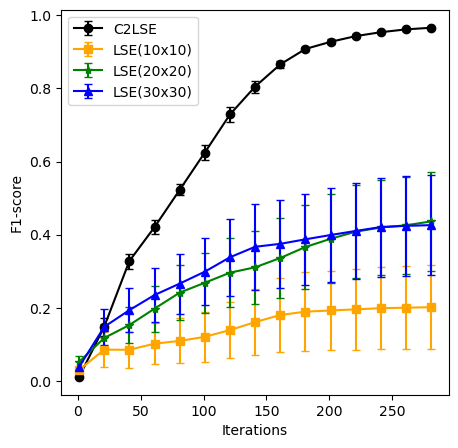}
        }
        \subfigure[SIN2D]{%
            \label{fig:gridsize:sin2d}
            \includegraphics[width=4.85cm]{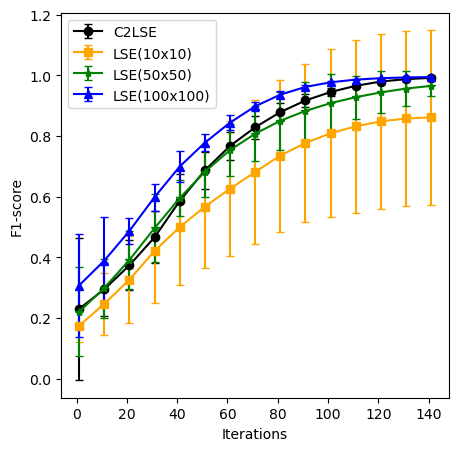}
        }
    }
\end{figure*}

\begin{figure*}[t]
    \floatconts
    {fig:gridsizecomp}
    {\caption{Comparing the computational cost of C2LSE with the LSE algorithm on \textit{different levels of discretization}.}}
    {%
        \subfigure[MC2D]{%
            \label{fig:gridsizecomp:mc2d}
            \includegraphics[width=4.85cm]{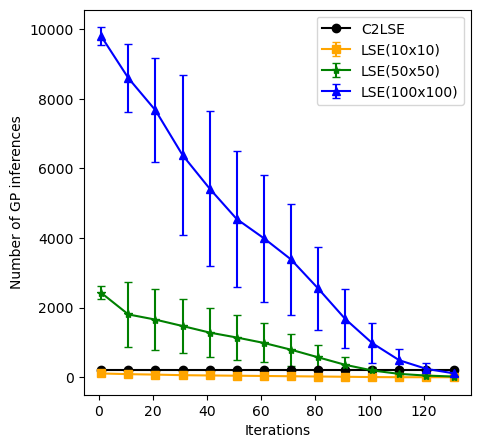}
        }\
        \subfigure[MC3D]{%
            \label{fig:gridsizecomp:mc3d}
            \includegraphics[width=4.85cm]{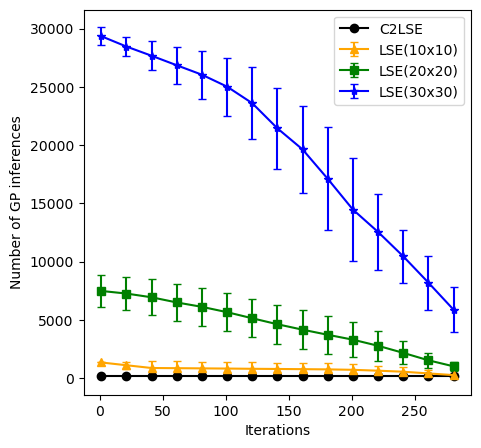}
        }
        \subfigure[SIN2D]{%
            \label{fig:gridsizecomp:sin2d}
            \includegraphics[width=4.85cm]{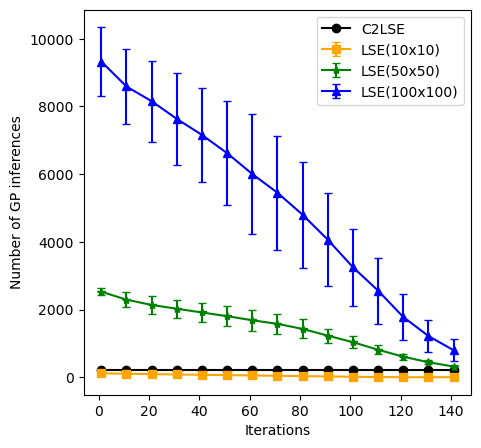}
        }
    }
\end{figure*}

On all three datasets, the LSE algorithm becomes less accurate when the input grid is more sparse. On the other hand, C2LSE does not depend on the resolution of the input set as it can be operated on a continuous search space. To achieve accurate classification, the LSE algorithm requires a densely superimposed grid (i.e. increasing the size of the finite search space). As a result, the number of GP inferences required to compute the acquisition function's values for all points in the grid is significantly larger than for C2LSE and for more coarse grids as can be seen in Figure \ref{fig:gridsizecomp}.

\subsubsection{Ablation study}

\begin{figure*}[t]
    \floatconts
    {fig:ablation}
    {\caption{Effect of \(\epsilon\) on the result of C2LSE}}
    {%
        \subfigure[MC2D]{%
            \label{fig:ablation:MC2D}
            \includegraphics[width=4.85cm]{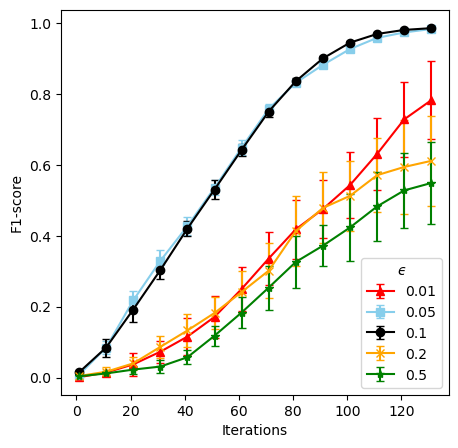}
        }
        \subfigure[MC3D]{%
            \label{fig:ablation:MC3D}
            \includegraphics[width=4.85cm]{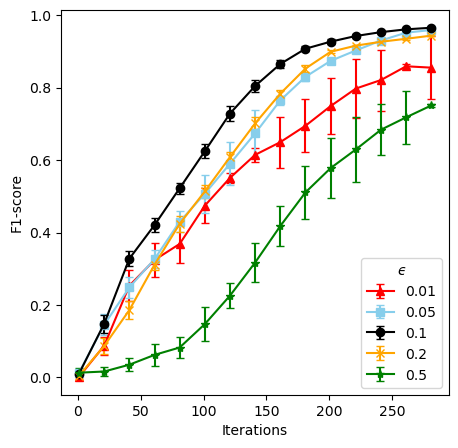}
        }
        \subfigure[SIN2D]{%
            \label{fig:ablation:SIN2D}
            \includegraphics[width=4.85cm]{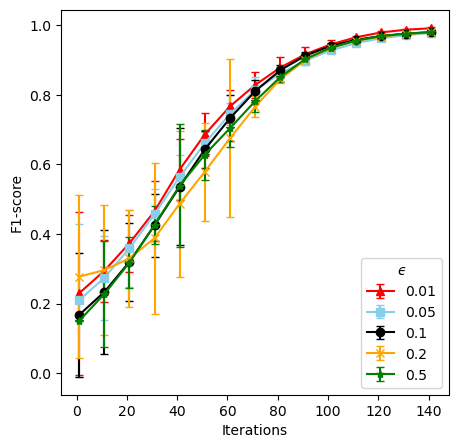}
        }
    }
\end{figure*}

\begin{figure*}[t]
    \floatconts
    {fig:ablation2}
    {\caption{Effect of \(\epsilon\) on the sampling behavior of C2LSE on MC2D. Again, the superlevel set is shown in white and the sublevel set is in gray. The black dots show the points queried by C2LSE after 20 iterations.}}
    {%
        \subfigure[\(\epsilon=0.01\)]{%
            \label{fig:MC2D0.01}
            \includegraphics[width=4.4cm]{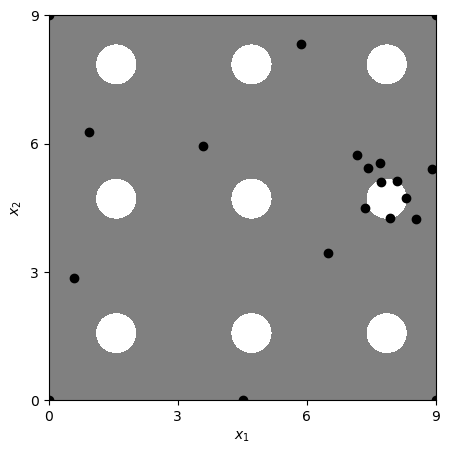}
        }
        \subfigure[\(\epsilon=0.1\)]{%
            \label{fig:MC2D0.1}
            \includegraphics[width=4.4cm]{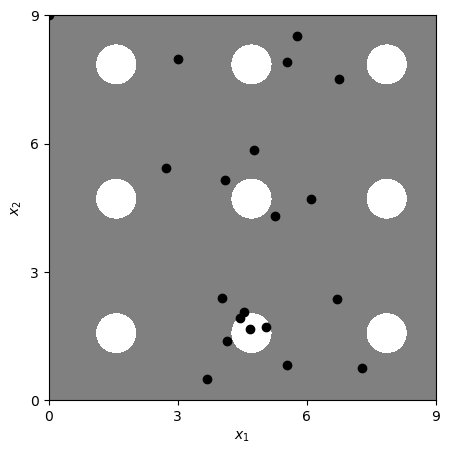}
        }
        \subfigure[\(\epsilon=0.5\)]{%
            \label{fig:MC2D0.5}
            \includegraphics[width=4.4cm]{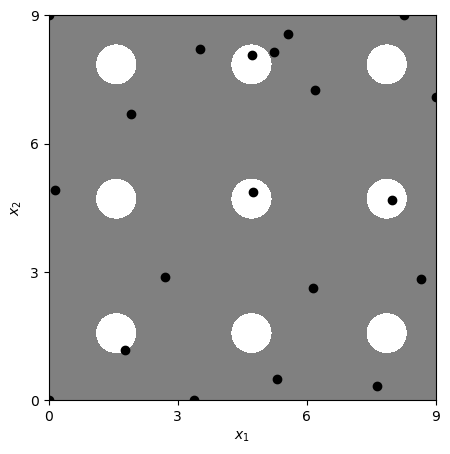}
        }
    }
\end{figure*}

We also study the impact of different values for \(\epsilon\) on the performance of C2LSE in this section. C2LSE is run with different values for \(\epsilon\) with the classification performance shown in Figure \ref{fig:ablation}. For large \(\epsilon\) values, the focus is on global exploration with queried points far away from the threshold boundary as can be seen in Figure \ref{fig:ablation2}c when \(\epsilon=0.5\). For small \(\epsilon\) values, the focus is on exploitation where points closest to the boundary are preferred as can be seen in Figure \ref{fig:ablation2}a. While classification will be highly accurate near the discovered regions in the case of good exploitation, the algorithm may not sufficiently explore when multiple disconnected superlevel/sublevel sets exist (e.g. the case of MC2D). Both exploration and exploitation are balanced when \(\epsilon=0.1\) with one region discovered and two more queried nearby as can be seen in Figure \ref{fig:ablation2}b.

\section{Conclusion}
To tackle the problem of active LSE on a continuous search space, we proposed C2LSE with the goal of suggesting new data points that are informative to the level set classification at each time step. Our algorithm aims to query where the current model is the least confident in classifying whether the point belongs to the superlevel or the sublevel sets. This confidence is reflected in our acquisition function which requires minimal computational cost. We theoretically analyzed the convergence rate of our algorithm showing its behavior in classifying any point in the continuous search space with not only high confidence but also high accuracy. The superior performance of C2LSE is demonstrated through several numerical experiments on both synthetic and real-world datasets with small associated computational costs.

\acks{This research was partially supported by the Australian Government through the Australian Research Council's Discovery Projects funding scheme (project DP210102798). The views expressed herein are those of the authors and are not necessarily those of the Australian Government or Australian Research Council.}

\bibliography{acml23}

\begin{thebibliography}{15}
\providecommand{\natexlab}[1]{#1}
\providecommand{\url}[1]{\texttt{#1}}
\expandafter\ifx\csname urlstyle\endcsname\relax
  \providecommand{\doi}[1]{doi: #1}\else
  \providecommand{\doi}{doi: \begingroup \urlstyle{rm}\Url}\fi

\bibitem[Bogunovic et~al.(2016)Bogunovic, Scarlett, Krause, and Cevher]{truvar}
Ilija Bogunovic, Jonathan Scarlett, Andreas Krause, and Volkan Cevher.
\newblock Truncated variance reduction: A unified approach to bayesian optimization and level-set estimation.
\newblock In \emph{Proceedings of the 30th International Conference on Neural Information Processing Systems}, NIPS'16, page 1515–1523, Red Hook, NY, USA, 2016. Curran Associates Inc.

\bibitem[Bryan et~al.(2005)Bryan, Schneider, Nichol, Miller, Genovese, and Wasserman]{straddle}
Brent Bryan, Jeff Schneider, Robert~C. Nichol, Christopher~J. Miller, Christopher~R. Genovese, and Larry Wasserman.
\newblock Active learning for identifying function threshold boundaries.
\newblock In \emph{Proceedings of the 18th International Conference on Neural Information Processing Systems}, NIPS'05, page 163–170, Cambridge, MA, USA, 2005. MIT Press.

\bibitem[Cover and Thomas(2006)]{10.5555/1146355}
Thomas~M. Cover and Joy~A. Thomas.
\newblock \emph{Elements of Information Theory (Wiley Series in Telecommunications and Signal Processing)}.
\newblock Wiley-Interscience, USA, 2006.

\bibitem[Gopakumar et~al.(2018)Gopakumar, Gupta, Rana, Nguyen, and Venkatesh]{NEURIPS2018_cc709032}
Shivapratap Gopakumar, Sunil Gupta, Santu Rana, Vu~Nguyen, and Svetha Venkatesh.
\newblock Algorithmic assurance: An active approach to algorithmic testing using bayesian optimisation.
\newblock In S.~Bengio, H.~Wallach, H.~Larochelle, K.~Grauman, N.~Cesa-Bianchi, and R.~Garnett, editors, \emph{Advances in Neural Information Processing Systems}, volume~31. Curran Associates, Inc., 2018.

\bibitem[Gotovos et~al.(2013)Gotovos, Casati, Hitz, and Krause]{activelse}
Alkis Gotovos, Nathalie Casati, Gregory Hitz, and Andreas Krause.
\newblock Active learning for level set estimation.
\newblock In \emph{Proceedings of the Twenty-Third International Joint Conference on Artificial Intelligence}, IJCAI '13, page 1344–1350. AAAI Press, 2013.

\bibitem[Ha et~al.(2021)Ha, Gupta, Rana, and Venkatesh]{HaHuong2021HDLS}
Huong Ha, Sunil Gupta, Santu Rana, and Svetha Venkatesh.
\newblock High dimensional level set estimation with bayesian neural network.
\newblock In \emph{Proceedings of the AAAI Conference on Artificial Intelligence}, volume~35, pages 12095--12103, 2021.

\bibitem[Hallquist and Manual(1998)]{hallquist1998livermore}
John~O Hallquist and LS-DYNA~Theoretical Manual.
\newblock Livermore software technology corporation.
\newblock \emph{Livermore, Ca}, pages 94550--1740, 1998.

\bibitem[Inatsu et~al.(2020)Inatsu, Karasuyama, Inoue, and Takeuchi]{9272630}
Yu~Inatsu, Masayuki Karasuyama, Keiichi Inoue, and Ichiro Takeuchi.
\newblock Active learning for level set estimation under input uncertainty and its extensions.
\newblock \emph{Neural Computation}, 32\penalty0 (12):\penalty0 2486--2531, 2020.

\bibitem[Inatsu et~al.(2021)Inatsu, Iwazaki, and Takeuchi]{inatsu2021active}
Yu~Inatsu, Shogo Iwazaki, and Ichiro Takeuchi.
\newblock Active learning for distributionally robust level-set estimation.
\newblock In \emph{International Conference on Machine Learning}, pages 4574--4584. PMLR, 2021.

\bibitem[Iwazaki et~al.(2020)Iwazaki, Inatsu, and Takeuchi]{9252941}
Shogo Iwazaki, Yu~Inatsu, and Ichiro Takeuchi.
\newblock Bayesian experimental design for finding reliable level set under input uncertainty.
\newblock \emph{IEEE Access}, 8:\penalty0 203982--203993, 2020.

\bibitem[Lecun et~al.(1998)Lecun, Bottou, Bengio, and Haffner]{lenet}
Yann Lecun, Leon Bottou, Y.~Bengio, and Patrick Haffner.
\newblock Gradient-based learning applied to document recognition.
\newblock \emph{Proceedings of the IEEE}, 86:\penalty0 2278 -- 2324, 12 1998.

\bibitem[Rasmussen and Williams(2005)]{10.7551/mitpress/3206.001.0001}
Carl~Edward Rasmussen and Christopher K.~I. Williams.
\newblock \emph{{Gaussian Processes for Machine Learning}}.
\newblock The MIT Press, 11 2005.

\bibitem[Senadeera et~al.(2020)Senadeera, Rana, Gupta, and Venkatesh]{warping}
Manisha Senadeera, Santu Rana, Sunil Gupta, and Svetha Venkatesh.
\newblock Level set estimation with search space warping.
\newblock In \emph{Advances in Knowledge Discovery and Data Mining}, pages 827--839, Cham, 2020. Springer International Publishing.

\bibitem[Srinivas et~al.(2009)Srinivas, Krause, Kakade, and Seeger]{srinivas}
Niranjan Srinivas, Andreas Krause, Sham Kakade, and Matthias Seeger.
\newblock Gaussian process bandits without regret: An experimental design approach.
\newblock \emph{CoRR}, abs/0912.3995, 01 2009.

\bibitem[Zanette et~al.(2019)Zanette, Zhang, and Kochenderfer]{rmile}
Andrea Zanette, Junzi Zhang, and Mykel~J. Kochenderfer.
\newblock Robust super-level set estimation using gaussian processes.
\newblock In \emph{Machine Learning and Knowledge Discovery in Databases}, pages 276--291, Cham, 2019. Springer International Publishing.

\end{thebibliography}
\end{document}